\documentclass[final]{IEEEphot}

\usepackage{amsmath}
\usepackage{amssymb}
\usepackage{color}
\usepackage{array}
\usepackage{booktabs}
\usepackage{multirow}
\usepackage[numbers, sort&compress]{natbib}
\usepackage[
 bookmarks=true,bookmarksnumbered=true,bookmarksopen=true,bookmarksopenlevel=1,
 breaklinks=false,pdfborder={0 0 0},pdfborderstyle={},backref=false,colorlinks=true]{hyperref}
\usepackage{t1enc}
\usepackage{subfig}

\begin{document}

\title{Incident Light Frequency-based Image Defogging Algorithm}

\author{Wenbo~Zhang, Xiaorong~Hou}

\affil{School of Energy Science and Engineering, University of Electronic
Science and Technology of China, Chengdu 611731, China}

\maketitle



\begin{abstract}
Considering the problem of color distortion caused by the defogging algorithm based on dark channel prior, an improved algorithm was proposed to calculate the transmittance of all channels respectively. First, incident light frequency's effect on the transmittance of various color channels was analyzed according to the Beer-Lambert's Law, from which a proportion among various channel transmittances was derived; afterwards, images were preprocessed by down-sampling to refine transmittance, and then the original size was restored to enhance the operational efficiency of the algorithm; finally, the transmittance of all color channels was acquired in accordance with the proportion, and then the corresponding transmittance was used for image restoration in each channel. The experimental results show that compared with the existing algorithm, this improved image defogging algorithm could make image colors more natural, solve the problem of slightly higher color saturation caused by the existing algorithm, and shorten the operation time by four to nine times.
\end{abstract}


\section{Introduction}

The images photographed by an imaging device in a foggy environment
are born with poor visibility and low contrast, which have direct
influence on the safety of aviation, maritime transport and road traffic.
Moreover, various outdoor monitoring systems, such as video surveillance
system, cannot work reliably in bad weather. Therefore, simple and
efficient image defogging is a research subject with an important
practical value that can help improve the reliability and robustness
of visual systems.

Image defogging technology is an emerging hotspot of research on digital
image processing, on which foreign scholars \cite{choi2015referenceless,zhu2015fast,zhao2015single,wang2014single,tang2014robust,tang2016robust,zhang2015research}
have made extensive studies and achieved a series of theoretical and
application results. In summary, image defogging methods fall broadly
into two categories \cite{guofan2010Review}:
image enhancement-based defogging and physical imaging model-based
defogging.

The image enhancement-based defogging method does not consider the
cause of image degradation in foggy weather, but merely enhances foggy
images in line with their characteristics such as vague detail and
low contrast. This can weaken the fog effect on images, improve the
visibility of scenes, and enhance the contrast of images. In this
image enhancement-based method, that which is the most common is histogram
equalization, which can effectively enhance the contrast of images,
but owing to the uneven depth of scenes in foggy images, namely different
scenes are affected by fog in varying degree, global histogram equalization
cannot fully remove the fog effect, while some details are still vague.
In Literature \cite{zhupei2004animage},
the sky is first separated from the rest by local histogram equalization,
and then depth information matching is realized skillfully in the
non-sky zone by a moving template. This algorithm overcomes the defect
of global histogram equalization that processes details ineffectively
and avoids the effect of sky noise. But when this algorithm is applied,
sub-image selection easily leads to a block effect, and thereby cannot
improve the visual effect considerably.

The physical imaging model-based defogging method analyzes and processes
the degradation process of foggy images in depth. It first analyzes
the inverse generation process of degraded images, and builds a model
for atmospheric scattering effect on the attenuation of image contrast
in order to restore the definition of degraded images. Generally speaking,
since this physical imaging model-based method considers the physical
degradation process of foggy images, images can be enhanced more effectively
in this way than only image processing is considered. Y. Yitzhaky
et al. \cite{yitzhaky1997restoration} were the first to consider
the cause of foggy image degradation. Based on a detailed analysis
of atmospheric effect on image degradation, they built an image degradation
model. If atmospheric effect on image degradation is considered to
be a process in which images go through a degradation system, an image
degradation model can be built to eliminate the influence of weather
factors on image quality. However, since the key to building an image
degradation model is to determine an atmospheric modulation transfer
function, and that the ratio of the effect of atmospheric turbulence
and aerosol particles in the atmospheric modulation transfer function
has something to do with the meteorological conditions of image photographing,
the corresponding local meteorological parameters should be acquired
from the meteorological station. However, these parameters are usually
hard to get due to the harsh additional conditions.

Recently, He \cite{he2011single}{]} et al. proposed a simple and
effective dark channel prior-based single image defogging method based
on the statistical law of outdoor fogless image databases. This method
can defog most outdoor images effectively. But in practical application,
this algorithm inevitably causes image distortion since excessive
color saturation is unavoidable. For this reason, this paper proposed
an improved algorithm. First, incident light frequency’s effect on
the transmittance of various color channels was analyzed according
to the Beer-Lambert’s Law, from which a proportion among various channel
transmittances was derived; after that, images were preprocessed by
down-sampling to refine transmittance, and then the original size
was restored to enhance the operational efficiency of the algorithm;
finally, the transmittance of all color channels was acquired in accordance
with the proportion, and then the corresponding transmittance was
used for image restoration in each channel.

The rest of this paper has the following structure: Section \ref{sec:2}
outlines the principle of the defogging algorithm based on dark channel
prior; Section \ref{sec:3} analyzes the deficiencies of the original
algorithm, and derives an improved algorithm; Section \ref{sec:4}
proves the validity of this improved algorithm by comparing it with
the existing algorithm in experiments.

\section{Dark Channel Prior-based Defogging Algorithm\label{sec:2}}

For a clear description below and an effective comparison in experiments,
this section outlines some major dark channel prior-based defogging
methods.

\subsection{Atmospheric Scattering Model}

The atmospheric scattering model proposed by Narasimhan et al. \cite{narasimhan2003contrast,narasimhan2002vision,schechner2001instant,cozman1997depth,nayar1999vision}
describes the degradation process of foggy images.

\begin{equation}
I(x)=J(x)t(x)+A(1-t(x))\label{eq:1}
\end{equation}

Where $I$ represents the intensity of the image observed, $J$ represents
the intensity of scene light, $A$ represents the atmospheric light
at infinity, and $t$ is known as transmittance. The first equation
item $J(x)t(x)$ is an attenuation term, and $A\left(1-t\left(x\right)\right)$
is an atmospheric light item. The aim of image defogging is to restore
$J$ from $I$ .

\subsection{Dark Channel Prior}

Dark channel prior knowledge comes from statistical observations of
a great many outdoor fogless images. It shows that there are always
some pixels in the overwhelming majority of images that have a very
small value in a color channel. This prior knowledge can be defined
as follows:

\[
J^{dark}\left(x\right)=\underset{c\in\left\{ r,g,b\right\} }{\min}\left(\underset{y\in\Omega\left(x\right)}{\min}\left(J^{c}\left(y\right)\right)\right)
\]

$J^{c}$ represents a color channel of $J$ , while $\Omega\left(x\right)$
is a pixel $x$ –centered square area. Suppose $J$ is an outdoor
fogless image, and $J^{dark}$ is a dark channel of $J$ , and the
above experiential law obtained by observations is known as dark channel
prior. Dark channel prior knowledge indicates that the value of $J^{dark}$
is always very low and close to 0.

\subsection{Defogging by Dark Channel Prior}

Suppose that atmospheric light $A$ has been fixed; then suppose transmittance
is constant in a local region. The minimum operator is adopted for
Equation (\ref{eq:1}), and meanwhile $A$ is divided, reducing to:

\[
\underset{y\in\Omega\left(x\right)}{\mathop{\min}}\,\left(\frac{{{I}^{c}}\left(y\right)}{{{A}^{c}}}\right)=\tilde{t}\left(x\right)\underset{y\in\Omega\left(x\right)}{\mathop{\min}}\,\left(\frac{{{J}^{c}}\left(y\right)}{{{A}^{c}}}\right)+\left(1-\tilde{t}\left(x\right)\right)
\]

Where superscript $c$ denotes the component of a certain color channel,
and $\tilde{t}(x)$ denotes a roughly estimated transmittance. The
minimum operator is adopted for color channel $c$, so,

\begin{equation}
\underset{c}{\mathop{\min}}\,\left(\underset{y\in\Omega\left(x\right)}{\mathop{\min}}\,\left(\frac{{{I}^{c}}\left(y\right)}{{{A}^{c}}}\right)\right)=\tilde{t}\left(x\right)\underset{c}{\mathop{\min}}\,\left(\underset{y\in\Omega\left(x\right)}{\mathop{\min}}\,\left(\frac{{{J}^{c}}\left(y\right)}{{{A}^{c}}}\right)\right)+\left(1-\tilde{t}\left(x\right)\right)\label{eq:2}
\end{equation}

According to the law of dark channel prior, the dark channel item
$J^{dark}$ in the outdoor fogless images should approach 0:

\[
{{J}^{dark}}\left(x\right)=\underset{c}{\mathop{\min}}\,\left(\underset{y\in\Omega\left(x\right)}{\mathop{\min}}\,\left({{J}^{c}}\left(y\right)\right)\right)=0
\]

A rough transmittance can be estimated if the above equation is substituted
into (\ref{eq:2}):

\begin{equation}
\tilde{t}\left(x\right)=1-\underset{c}{\mathop{\min}}\,\left(\underset{y\in\Omega\left(x\right)}{\mathop{\min}}\,\left(\frac{{{I}^{c}}\left(y\right)}{{{A}^{c}}}\right)\right)\label{eq:3}
\end{equation}

It has been discovered in practical application that if fog is removed
thoroughly, an image will, however, look unreal, and depth perception
will be lost. Therefore, constant $\omega\left(0<\omega\le1\right)$
can be introduced to equation (\ref{eq:3})to retain some fog:

\[
\tilde{t}\left(x\right)=1-\underset{c}{\mathop{\omega\min}}\,\left(\underset{y\in\Omega\left(x\right)}{\mathop{\min}}\,\left(\frac{{{I}^{c}}\left(y\right)}{{{A}^{c}}}\right)\right)
\]

Transmittance can only be roughly estimated according to the above
equation, so to improve the accuracy, the original paper used an image
matting algorithm\cite{levin2008closed}to refine the transmittance.
The following linear equation can be solved to refine the transmittance:

\begin{equation}
\left(L+\lambda U\right)t=\lambda\tilde{t}\label{eq:4}
\end{equation}

where $\lambda$ is a corrected parameter, $L$ is the Laplacian matrix
proposed by the image matting algorithm, which is usually a large
sparse matrix.

After a refined transmittance $t\left(x\right)$ is obtained, the
equation below is used to calculate the resulting image $J\left(x\right)$
of defogging:

\begin{equation}
J\left(x\right)=\frac{I\left(x\right)-A}{\max\left(t\left(x\right),{{t}_{0}}\right)}+A\label{eq:5}
\end{equation}

Where atmospheric light $A$ is estimated this way: sort the pixels
in dark channel $J^{dark}$ in descending order in accordance with
brightness value, compare the brightness value of the first 0.1\%
pixels with their brightness value in the original image $I$, and
finally take the brightest point as atmospheric light $A$.

For most outdoor foggy images, the above algorithm can achieve a good
defogging effect, but color distortion or excessive color saturation
may be caused when this algorithm is used to process some images.
Moreover, the algorithm runs very slowly. For instance, it will take
38.3 seconds to input an image of 600{*}455. To solve this problem,
this paper improved the original algorithm.

\section{Incident Light Frequency-based Algorithm\label{sec:3}}

The imaging formula popular in the machine vision field is adopted
in the existing defogging models\cite{nayar1999vision,narasimhan2002vision,cozman1997depth,narasimhan2003contrast,schechner2001instant},
which was derived based on the Beer-Lambert Law\cite{swinehart1962beer}:

\begin{equation}
I(x)={{L}_{\infty}}\rho(x){{e}^{-\beta d(x)}}+{{L}_{\infty}}(1-{{e}^{-\beta d(x)}})\label{eq:6}
\end{equation}

where $I(x)$ represents the brightness value of the pixel at $x$
coordinate in the image, $\rho(x)$ represents the reflection coefficient
of various body surfaces, and $e^{-\beta d(x)}$ is transmittance
$t$, which represents the attenuation degree of energy when light
propagates in atmosphere.

As can be seen from the proof procedure of the Beer-Lambert Law, transmittance
is derived from the equation below:

\begin{equation}
t={{e}^{-\int_{{{Z}_{0}}}^{{{Z}_{1}}}{\gamma(\nu,z)dz}}}\label{eq:7}
\end{equation}

where $\nu$ represents incident light frequency, $z$ represents
a certain point on the propagation path of incident light. As shown
in (\ref{eq:7}), transmittance $t$ is related to the medium attribute
$\gamma\left(\nu,z\right)$ of each point on the propagation path
of incident light.

\subsection{Original Algorithm Hypotheses}

In order to reduce the complexity of the existing defogging model,
two hypotheses are made on transmittance $t$ in it.

\subsubsection{Suppose incident light has constant frequency}

When incident light frequency $\nu$ is constant, equation (\ref{eq:7})is
simplified into:

\begin{equation}
\begin{split}t & ={{e}^{-\int_{{{Z}_{0}}}^{{{Z}_{1}}}{\gamma(\nu,z)dz}}}\\
 & ={{e}^{-\int_{{{Z}_{0}}}^{{{Z}_{1}}}{\beta*D(z)dz}}}
\end{split}
\label{eq:8}
\end{equation}

As can be seen in the equation above, the medium attribute function
on the propagation path is simplified from bivariate function $\gamma\left(\nu,z\right)$
into single-variable function $D(z)$.

\subsubsection{Suppose there are homogeneous atmospheric media on the propagation
path of incident light}

Under this hypothesis, equation (\ref{eq:8}) is further simplified
into:

\begin{align*}
t & ={{e}^{-\int_{{{Z}_{0}}}^{{{Z}_{1}}}{\beta*D(z)dz}}}\\
 & ={{e}^{-\beta*\int_{{{Z}_{0}}}^{{{Z}_{1}}}{Ddz}}}\\
 & ={{e}^{-\beta d(x)}}
\end{align*}

Where $d(x)$ represents the field depth at point $x$ in the image,
namely the spatial distance between object and imaging device.

\subsection{Improvement Direction}

Although the above two hypotheses have greatly simplified the complexity
of the defogging model, defogging quality have been lowered significantly.
In order to further improve defogging quality, we reintroduced the
effect of incident light frequency $\nu$ on attenuation coefficient
$\beta$ into atmospheric light imaging formula (\ref{eq:6}), thus
further improving the atmospheric light imaging formula, as shown
below:

\begin{equation}
I(x)={{L}_{\infty}}\rho(x){{e}^{-\beta(\nu)d(x)}}+{{L}_{\infty}}(1-{{e}^{-\beta(\nu)d(x)}})\label{eq:9}
\end{equation}

At this point, attenuation coefficient $\beta$ is turned into the
function $\beta(v)$ of incident light frequency $v$.

According to the distance between object and imaging device (field
depth), foggy images can fall into three categories:
\begin{itemize}
\item Long-field images. The overwhelming majority of scenes in the images
are in the range of long field depth (objects are over 500m away from
the camera).
\item Short-field images. The overwhelming majority of scenes in the images
are in the range of short field depth (objects are less than 500m
away from the camera).
\item Mixed-field images. Long and short-field scenes exist side by side
in the images.
\end{itemize}
For long-field images, since field depth has increased, the concentration
of fog on the propagation path of incident light will show increasingly
complex changes with distance increasing. So, the prerequisites for
the tenability of Equation (\ref{eq:9}): media are no longer uniformly
distributed on the propagation path of light, and a new atmospheric
light model needs to be built. According to the observation of the
real world, if an observed object is farther away from the observer,
the light from it will be harder to discover and thereby be replaced
with atmospheric light (the sum of the various light beams from the
environment that interact with each other). Thus, the analytical processing
of such images can be translated into that of atmospheric light distribution.

For short-field images, due to the small field depth, the concentration
of fog within this range can be considered constant, so such foggy
images can be processed by Equation (\ref{eq:9}).

For mixed-field images, zones can be partitioned according to field
depth, and then images can be processed separately in accordance with
scene types.

This paper proposed an improved method for the processing of short-field
images. As can be seen from Equation (\ref{eq:9}), the key to the
processing of short-field images lies in the calculation of attenuation
coefficient $\beta(v)$. That is the focus of research in this paper.

\subsection{Attenuation Coefficient}

Since only the incident light from such three frequency bands as red
(R), Green (G) and Blue (B) is imaged by a sensing method in the current
cameras, the analysis of $\beta(v)$ can be limited to R, G and B.
A typical frequency value ${{v}_{r}},{{v}_{g}},{{v}_{b}}$ is fetched
respectively from R, G and B, which correspond to attenuation coefficient
$\beta({{v}_{r}}),\beta({{v}_{g}}),\beta({{v}_{b}})$ respectively.
The value of attenuation coefficient can be calculated by the analytical
statistics of the attenuation of pure light R, G and B in foggy weather.

After the value of $\text{\ensuremath{\beta}({\ensuremath{{v}_{r}}}),\ensuremath{\beta}({\ensuremath{{v}_{g}}}),\ensuremath{\beta}({\ensuremath{{v}_{b}}})}$
is calculated, their ratio can be computed. The result is denoted
by:

\[
\beta({{v}_{r}}):\beta({{v}_{g}}):\beta({{v}_{b}})={{\beta}_{r}}:{{\beta}_{g}}:{{\beta}_{b}}
\]

The statistical test shows that the effect of image restoration is
comparatively ideal when $\ensuremath{{{\beta}_{r}}:{{\beta}_{g}}:{{\beta}_{b}}=1:1.28:1.61}$.

Suppose it is known that some color channel’s transmittance $t_{c}=e^{-\beta_{c}d(x)}$,
other channels’ transmittance $\tilde{t}_{c}={e^{-{{\tilde{\beta}}_{c}}d(x)}}$
can be calculated by it:

\begin{equation}
\begin{split}{{\tilde{t}}_{c}} & ={{e}^{-{{\tilde{\beta}}_{c}}d(x)}}\\
 & ={{e}^{-\lambda{{\beta}_{c}}d(x)}}\\
 & ={{({{e}^{-{{\beta}_{c}}d(x)}})}^{\lambda}}\\
 & ={{t}_{c}}^{\lambda}
\end{split}
\end{equation}

Where $c\in\{r,g,b\},\lambda={{\tilde{\beta}}_{c}}/{{\beta}_{c}}$.

As can be seen in the equation above, once a color channel’s transmittance$t_{c}$
is worked out, other color channels’ transmittance can be calculated
according to it, as shown below:

\begin{equation}
{{t}_{c}}=\left\{ \begin{array}{ll}
1-\omega\underset{c}{\mathop{\min}}\,\left(\underset{y\in\Omega\left(x\right)}{\mathop{\min}}\,\left(\frac{{{I}^{c}}(y)}{{{A}^{c}}}\right)\right) & ;c=d,and\,d~is~dark~channel.\\
{{t}_{d}}^{\lambda} & ;c\ne d,then\,\lambda=\frac{{{\beta}_{c}}}{{{\beta}_{d}}}
\end{array}\right.
\end{equation}
\label{eq:11}

\subsection{An Improved Image Restoration Method}

Based on the analysis above, this section put forward a new transmittance
calculation method.

The main steps are shown as follows:
\begin{enumerate}
\item Down-sample output image $I\left(x\right)$, to reduce image size
to $\hat{I}\left(x\right)$.
\item Acquire dark channel pixel value ${{\hat{I}}_{d}}\left(x\right)$
and its color channel $d$ according to the formula below:
\[
{{\hat{I}}_{d}}\left(x\right)=\underset{c\in\left\{ r,g,b\right\} }{\mathop{\min}}\,\left(\underset{y\in\Omega\left(x\right)}{\mathop{\min}}\,\left({{\hat{I}}^{c}}\left(y\right)\right)\right)
\]
\item Calculate the transmittance ${{\tilde{t}}_{d}}\left(x\right)$ corresponding
to the color channel $d$ in which the dark channel is located using
the formula below:
\begin{equation}
{{\tilde{t}}_{d}}\left(x\right)=1-\omega{{\hat{I}}_{d}}\left(x\right)
\end{equation}
\label{eq:12}
\item Use image matting algorithm \cite{levin2008closed} namely solve Equation
(\ref{eq:4})to refine transmittance ${{\tilde{t}}_{d}}\left(x\right)$
into ${{\hat{t}}_{d}}\left(x\right)$.
\item Restore ${{\hat{t}}_{d}}\left(x\right)$ to the size of the original
image by interpolation and get ${{t}_{d}}\left(x\right)$.
\item Calculate the transmittance of all color channels by Equation (\ref{eq:11}).
\item Restore original image $J\left(x\right)$ in all color channels using
the equation below:
\[
{{J}_{c}}\left(x\right)=\frac{{{I}_{c}}\left(x\right)-{{A}_{c}}}{\max\left({{t}_{c}}\left(x\right),{{t}_{0}}\right)}+{{A}_{c}}\,\,\,;c\in\{r,g,b\}
\]
The rough operating process of the defogging algorithm is shown in
Fig.1.
\end{enumerate}
\begin{center}
\begin{figure}
\begin{centering}
\includegraphics[scale=0.8]{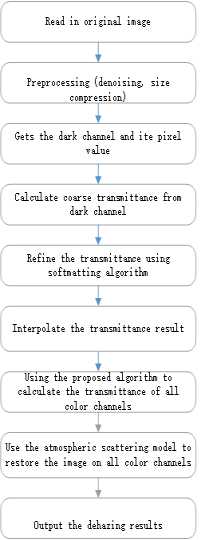}
\par\end{centering}
\caption{Algorithm processing flow chart.}

\end{figure}
\par\end{center}

\section{Experiments and Analysis\label{sec:4}}

To test the effect of this algorithm, this paper selected three images:
tiananmen of 600{*}455; house of 441{*}450; swan of 835{*}557.

The window $\Omega$ used for dark channel calculation has a size
of $5\times5$, and the $\omega$ in Equation (\ref{eq:12}) is set
equal to 0.95. The algorithm in Literature \cite{levin2008closed}
is still used for image matting.

For the testing experiment in this paper, we adopted the following
hardware platform: Intel(R) Core(TM) i3-3220 CPU with a basic frequency
of 3.3GHz; 8-G memory with a basic frequency of DDR3 1600MHz. A software
platform: MATLAB R2015b 64-bit. All algorithms were implemented through
MATLAB code.

Image matting was adopted to refine transmittance in the original
algorithm\cite{he2011single}. But this method essentially has high
time complexity and space complexity since it is usually used to solve
large-scale sparse linear equations. However, the effect of this step
on restoration is no more than softening the edge of the transition
region between foreground and background to weaken edge effect. Thus,
the algorithm proposed in this paper reduces image size significantly,
then refines transmittance by image matting, and finally restores
the refined transmittance image to the original size by tri-cubic
interpolates.

Fig.\ref{Fig:2} is a comparison of transmittance between the improved
algorithm and the original algorithm. As can be seen in the figure,
there is little difference between both in edge softening. However,
size reduction can help greatly improve the computational efficiency
of restoration algorithm. As can be seen in Tab.\ref{Tab:1}, the
operating efficiency of the algorithm proposed in this paper is 4\textasciitilde{}9
times as high as that of the original one.

According to the results before and after improvement shown in Fig.\ref{Fig:3}
to Fig.\ref{Fig:6}, after the transmittance of multiple channels
is corrected, the algorithm proposed in this paper solves the problem
of slightly higher color saturation in the existing algorithm, and
achieves a better and more natural visual effect than the original
algorithm.

\section{Conclusions}

This paper made a theoretical analysis and experimental observation
of the dark channel prior-based defogging algorithm, discovering from
its theoretical basis that the existing algorithm ignores the effect
of incident light frequency on transmittance. Therefore, this paper
started with the derivation process of transmittance to reversely
derive the relation between various channel transmittances to enhance
the defogging algorithm. The experimental results prove that this
improved algorithm can achieve a more natural color effect in the
restoration result. Since roughly calculated dark channel was still
used to compute the relationship between various transmittances, a
slight block effect appeared in the restoration result. The current
algorithm will be improved at the next step to eradicate this block
effect.
\begin{center}
\begin{figure}
\begin{centering}
\subfloat[]{\begin{centering}
\includegraphics[width=0.25\textwidth]{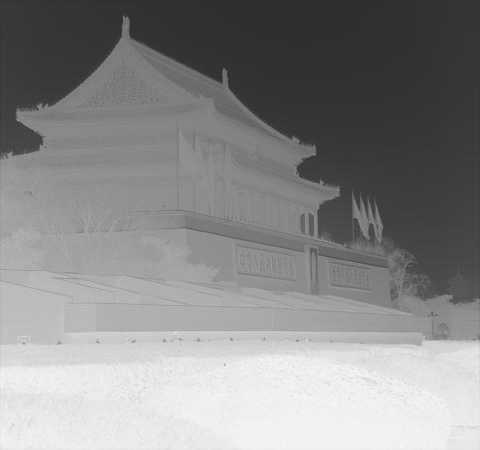}
\par\end{centering}
}\subfloat[]{\begin{centering}
\includegraphics[width=0.25\textwidth]{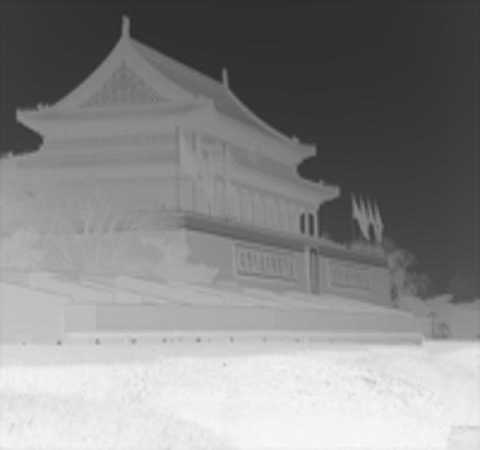}
\par\end{centering}}
\par\end{centering}
\begin{centering}
\subfloat[]{\centering{}\includegraphics[width=0.25\textwidth]{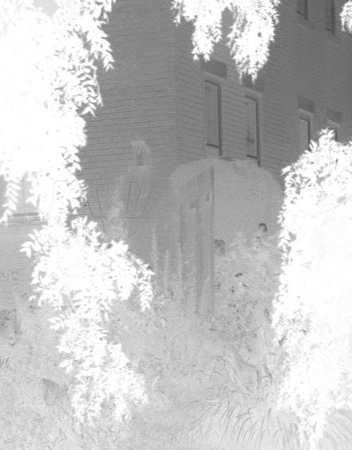}}\subfloat[]{\centering{}\includegraphics[width=0.25\textwidth]{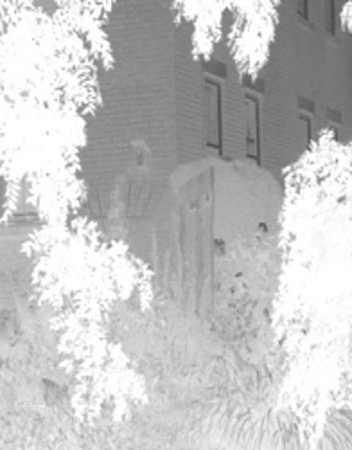}}
\par\end{centering}
\begin{centering}
\subfloat[]{\centering{}\includegraphics[width=0.25\textwidth]{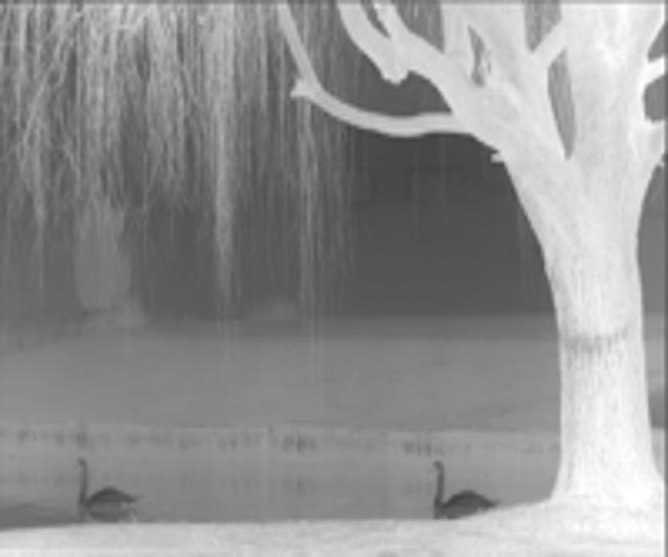}}\subfloat[]{\centering{}\includegraphics[width=0.25\textwidth]{Optimized_Transmission_Swan}}
\par\end{centering}
\caption{Comparison of the transmission maps.}
\label{Fig:2}
\end{figure}
\par\end{center}

\begin{center}
\begin{figure}
\subfloat[Original]{\centering{}\includegraphics[width=0.3\textwidth]{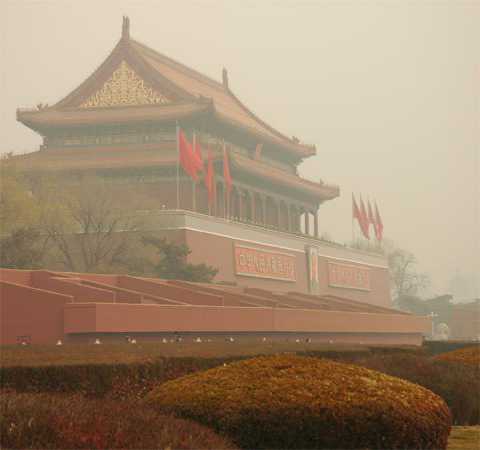}}\hspace*{\fill}\subfloat[He]{\begin{centering}
\includegraphics[width=0.3\textwidth]{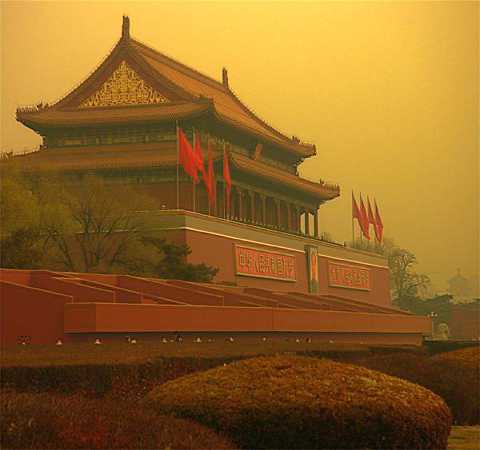}
\par\end{centering}
}\hspace*{\fill}\subfloat[This paper]{\begin{centering}
\includegraphics[width=0.3\textwidth]{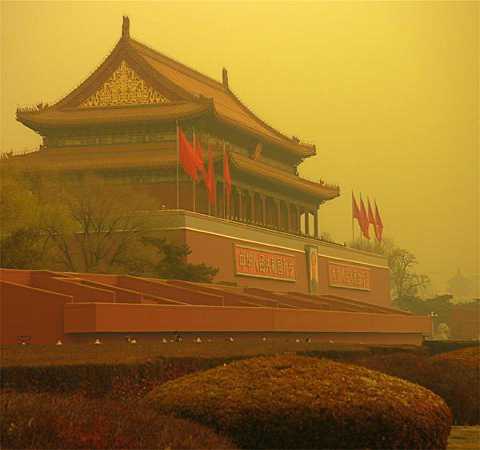}
\par\end{centering}

}\caption{Comparison of the dehazing results.}

\label{Fig:3}
\end{figure}
\par\end{center}

\begin{center}
\begin{figure}
\subfloat[Original]{\centering{}\includegraphics[width=0.3\textwidth]{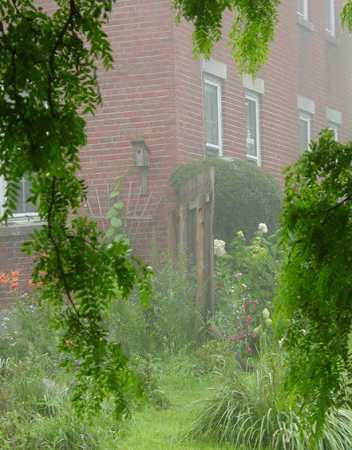}}\hspace*{\fill}\subfloat[He]{\begin{centering}
\includegraphics[width=0.3\textwidth]{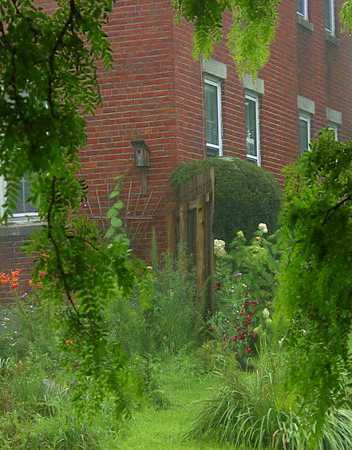}
\par\end{centering}
}\hspace*{\fill}\subfloat[This paper]{\begin{centering}
\includegraphics[width=0.3\textwidth]{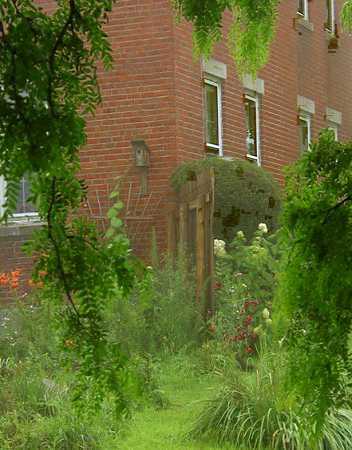}
\par\end{centering}
}\caption{Comparison of the dehazing results.}
\label{Fig:4}
\end{figure}
\par\end{center}

\begin{center}
\begin{figure}
\subfloat[Original]{\centering{}\includegraphics[width=0.3\textwidth]{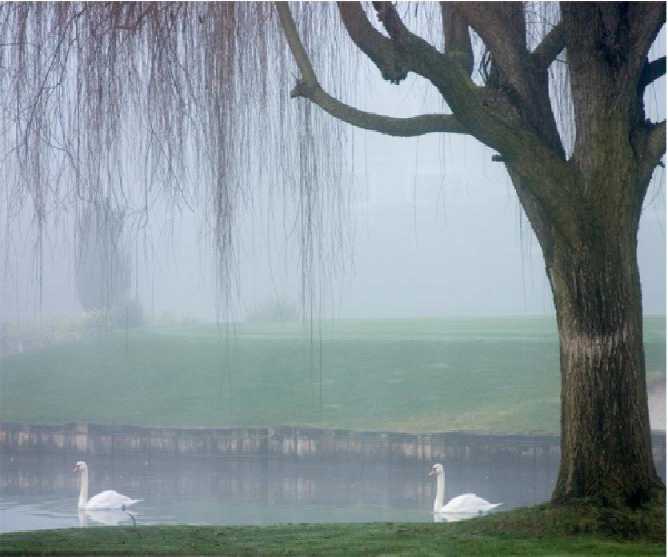}}\hspace*{\fill}\subfloat[He]{\begin{centering}
\includegraphics[width=0.3\textwidth]{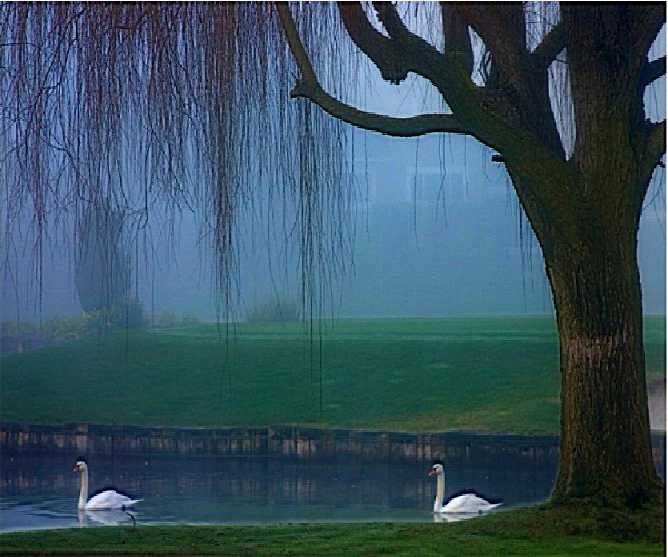}
\par\end{centering}
}\hspace*{\fill}\subfloat[This paper]{\begin{centering}
\includegraphics[width=0.3\textwidth]{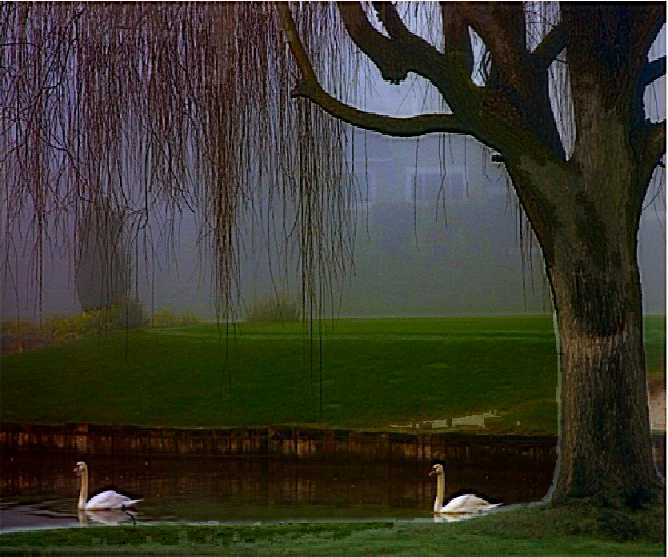}
\par\end{centering}
}\caption{Comparison of the dehazing results.}

\label{Fig:5}
\end{figure}
\par\end{center}

\begin{center}
\begin{figure}
\subfloat[Original]{\centering{}\includegraphics[width=0.3\textwidth]{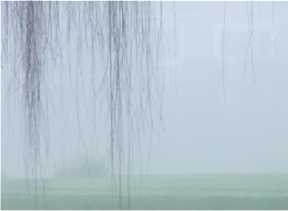}}\hspace*{\fill}\subfloat[He]{\begin{centering}
\includegraphics[width=0.3\textwidth]{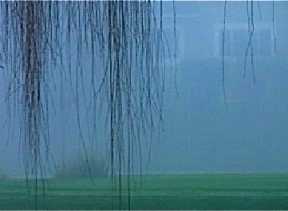}
\par\end{centering}
}\hspace*{\fill}\subfloat[This paper]{\begin{centering}
\includegraphics[width=0.3\textwidth]{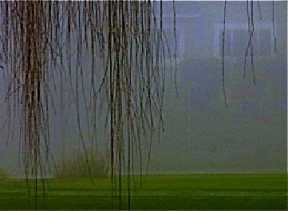}
\par\end{centering}
}\caption{Comparison of local amplification of the dehazing results.}

\label{Fig:6}
\end{figure}
\par\end{center}

\begin{center}
\begin{table}
\begin{centering}
\begin{tabular}{|c|c|c|c|c|}
\hline
Image name & Image size & He & This paper & Speed up\tabularnewline
\hline
\hline
tiananmen & 600{*}455 & 38.30s & 5.37s & 7.14\tabularnewline
\hline
house & 441{*}450 & 27.78s & 6.37s & 4.36\tabularnewline
\hline
swan & 835{*}557 & 70.95s & 7.22s & 9.82\tabularnewline
\hline
\end{tabular}
\par\end{centering}
\caption{Comparison of time consumption.}
\label{Tab:1}

\end{table}
\par\end{center}

\bibliographystyle{IEEEtran}
\bibliography{References}

\end{document}